\documentclass{article}
\usepackage{graphicx}

\usepackage{float}
\usepackage{wrapfig}
\usepackage{lipsum}
\usepackage{courier}
\usepackage{amsmath,amsfonts,amssymb,amsthm}
\usepackage[title,toc,titletoc,page]{appendix}

\usepackage[final,nonatbib]{neurips_2022}

\usepackage[utf8]{inputenc} 
\usepackage[T1]{fontenc}    
\usepackage{hyperref}       
\usepackage{url}            
\usepackage{booktabs}       
\usepackage{amsfonts}       
\usepackage{nicefrac}       
\usepackage{microtype}      
\usepackage{xcolor}         

\title{Personalizing Text-to-Image Generation via\\ Aesthetic Gradients}

\author{%
  Victor Gallego\\
  Komorebi AI Technologies\\
  \texttt{victor.gallego@komorebi.ai} 
}

\begin{document}

\maketitle
\begin{abstract}
This work proposes aesthetic gradients, a method to personalize a CLIP-conditioned diffusion model by guiding the generative process towards custom aesthetics defined by the user from a set of images. The approach is validated with qualitative and quantitative experiments, using the recent stable diffusion model and several aesthetically-filtered datasets. Code is released at \url{https://github.com/vicgalle/stable-diffusion-aesthetic-gradients}
\end{abstract}

\begin{figure}[!hbt]
\vspace{-20pt}
    \centering
    \setlength{\abovecaptionskip}{6.5pt}
    \setlength{\belowcaptionskip}{-3.5pt}
    \setlength{\tabcolsep}{0.55pt}
    \renewcommand{\arraystretch}{1.0}
    {
    
    \begin{tabular}{c}
    
    \begin{tabular}{c@{\hskip 5pt} c c @{\hskip 20pt} c c @{\hskip 20pt} c c @{\hskip 20pt} c c }
     \\

    \raisebox{0.045\linewidth}{\footnotesize\begin{tabular}{c@{}c@{}} Original \\ SD \end{tabular}} & 
    \includegraphics[width=0.09\linewidth,height=0.09\linewidth]{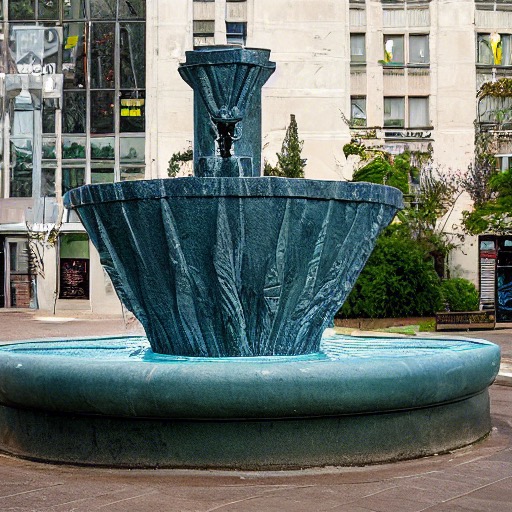} & 
    \includegraphics[width=0.09\linewidth,height=0.09\linewidth]{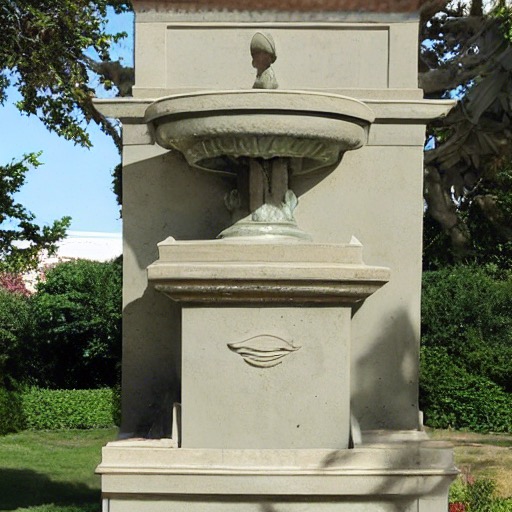} & 
    \includegraphics[width=0.09\linewidth,height=0.09\linewidth]{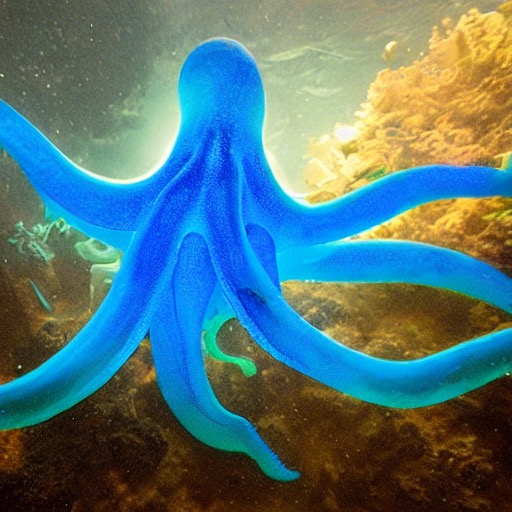} & 
    \includegraphics[width=0.09\linewidth,height=0.09\linewidth]{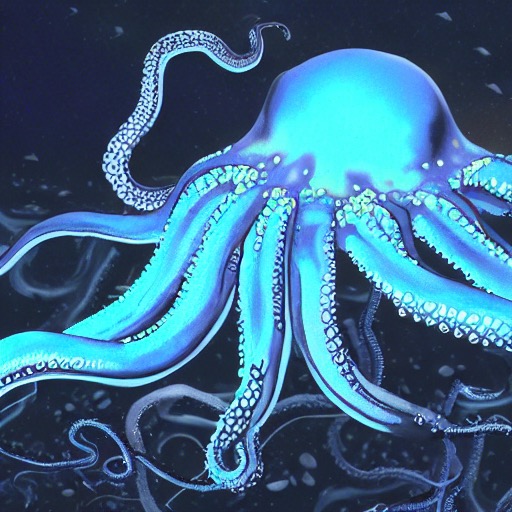} & 
    \includegraphics[width=0.09\linewidth,height=0.09\linewidth]{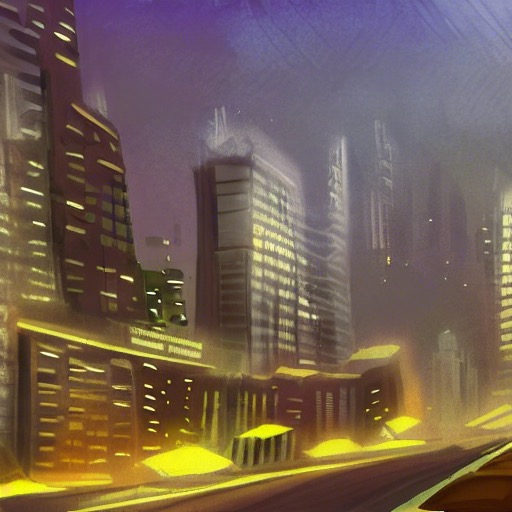} &
    \includegraphics[width=0.09\linewidth,height=0.09\linewidth]{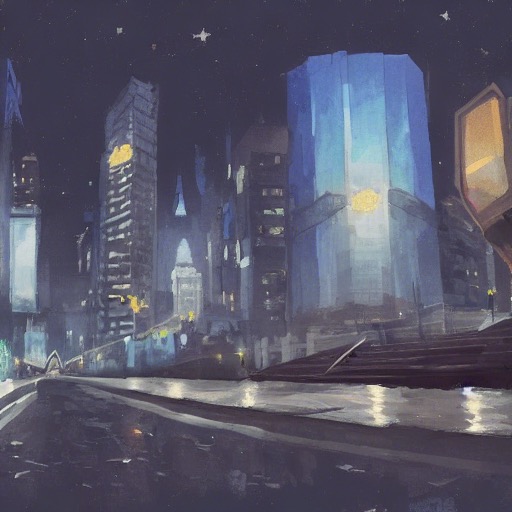} &
    \includegraphics[width=0.09\linewidth,height=0.09\linewidth]{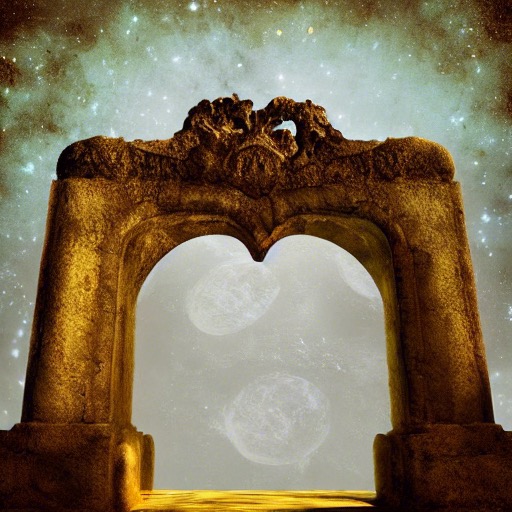} & 
    \includegraphics[width=0.09\linewidth,height=0.09\linewidth]{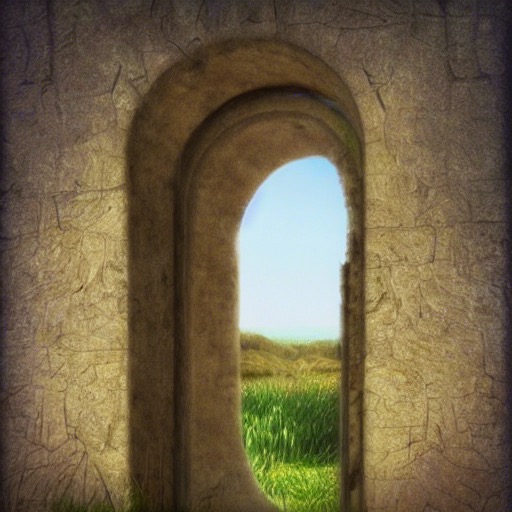} \\[2pt]
    
    \raisebox{0.045\linewidth}{\footnotesize\begin{tabular}{c@{}c@{}c@{}c@{}} Personalized \\ with $\mbox{SAC}_{8+}$ \end{tabular}}  &
    \includegraphics[width=0.09\linewidth,height=0.09\linewidth]{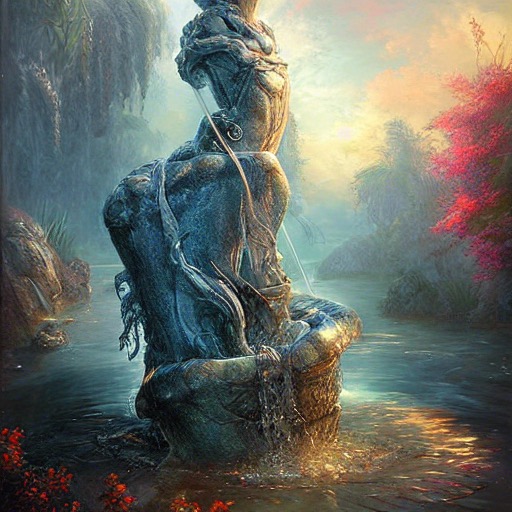} & 
    \includegraphics[width=0.09\linewidth,height=0.09\linewidth]{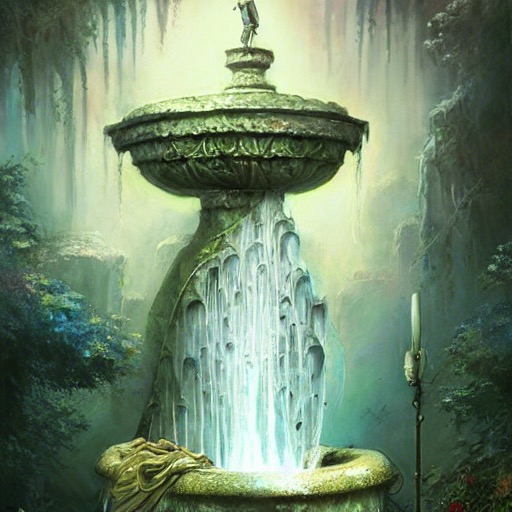} & 
    \includegraphics[width=0.09\linewidth,height=0.09\linewidth]{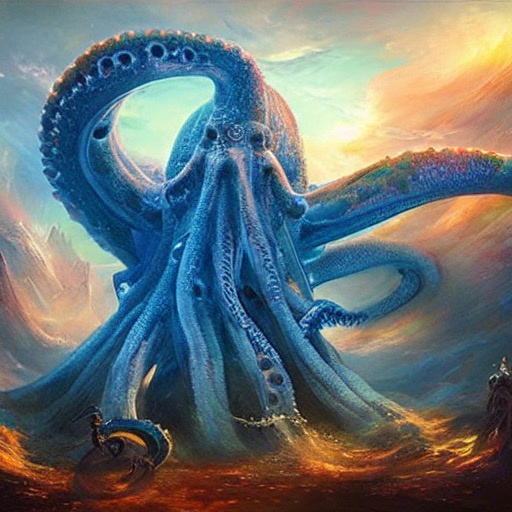} & 
    \includegraphics[width=0.09\linewidth,height=0.09\linewidth]{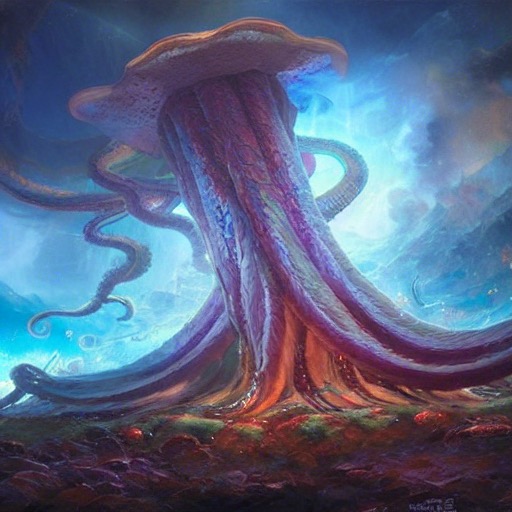} & 
    \includegraphics[width=0.09\linewidth,height=0.09\linewidth]{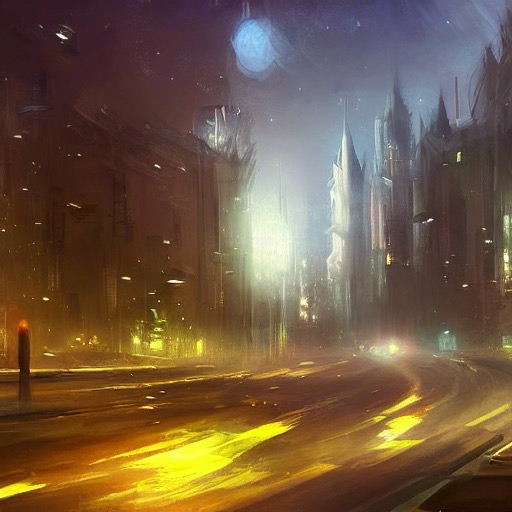} &
    \includegraphics[width=0.09\linewidth,height=0.09\linewidth]{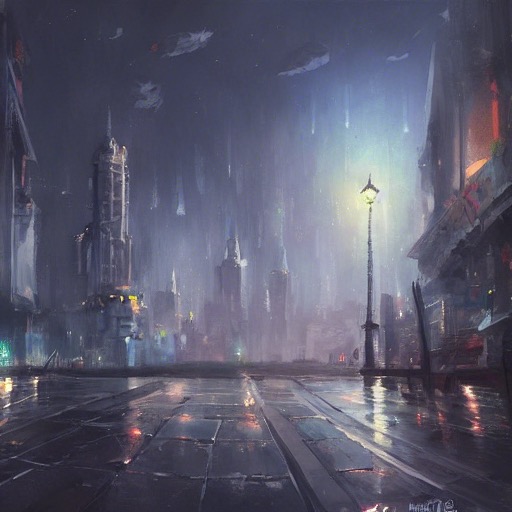} &
    \includegraphics[width=0.09\linewidth,height=0.09\linewidth]{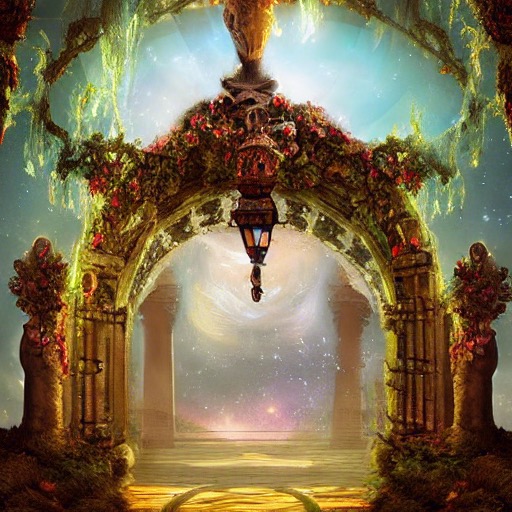} & 
    \includegraphics[width=0.09\linewidth,height=0.09\linewidth]{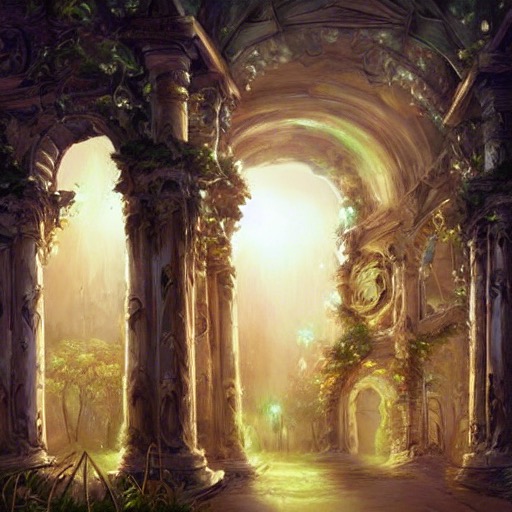} \\[-4pt]
        &
    \multicolumn{2}{l}{\tiny\begin{tabular}{c@{}c@{}c@{}} A fountain, sculpture \end{tabular}} &
    \multicolumn{2}{l}{\tiny\begin{tabular}{c@{}c@{}c@{}} A giant octopus, bioluminescence \end{tabular}} & 
    \multicolumn{2}{l}{\tiny\begin{tabular}{c@{}c@{}c@{}} A nighttime cityscape, concept art \end{tabular}} & 
    \multicolumn{2}{l}{\tiny\begin{tabular}{c@{}c@{}c@{}} A gateway between dreams \end{tabular}} \\[3pt]

     \raisebox{0.045\linewidth}{\footnotesize\begin{tabular}{c@{}c@{}} Original \\ SD \end{tabular}} & 
    \includegraphics[width=0.09\linewidth,height=0.09\linewidth]{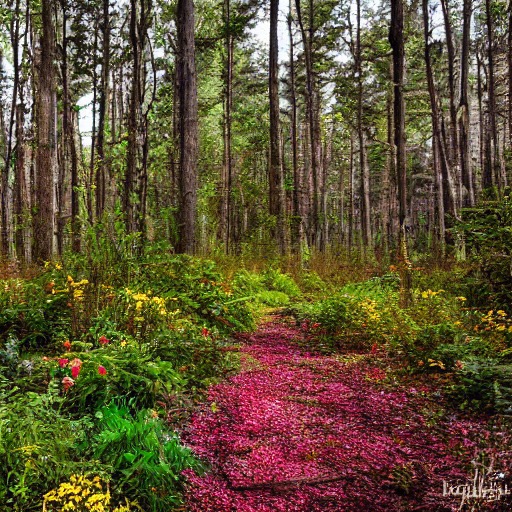} & 
    \includegraphics[width=0.09\linewidth,height=0.09\linewidth]{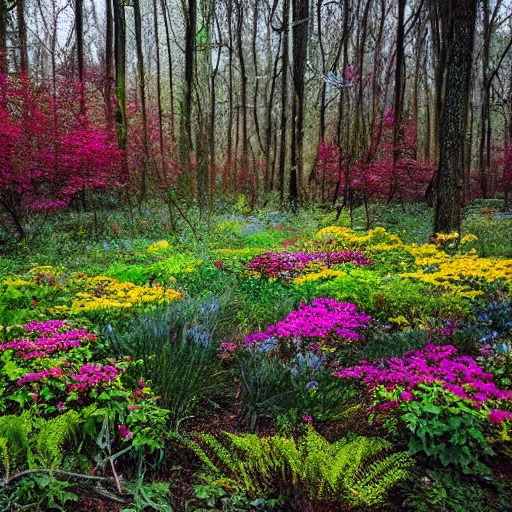} & 
    \includegraphics[width=0.09\linewidth,height=0.09\linewidth]{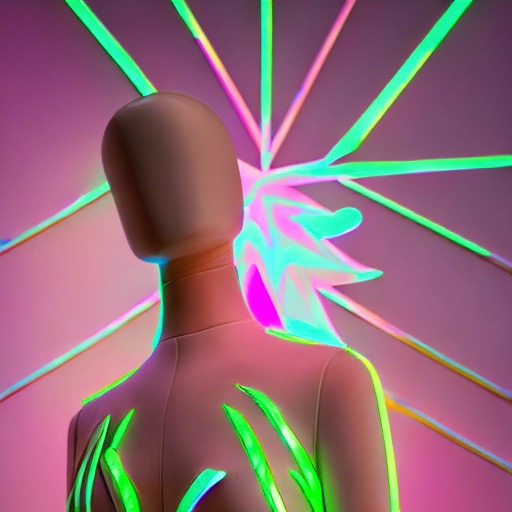} & 
    \includegraphics[width=0.09\linewidth,height=0.09\linewidth]{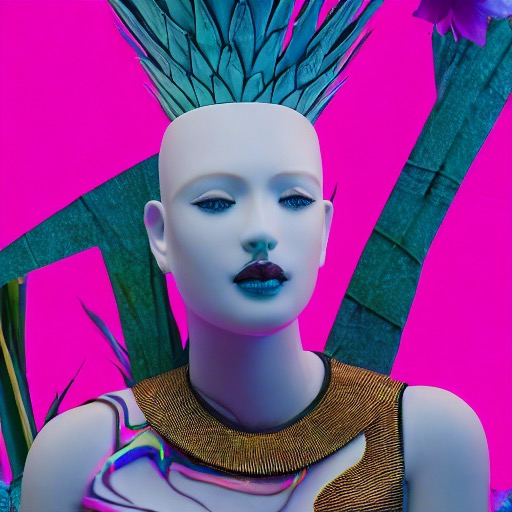} & 
    \includegraphics[width=0.09\linewidth,height=0.09\linewidth]{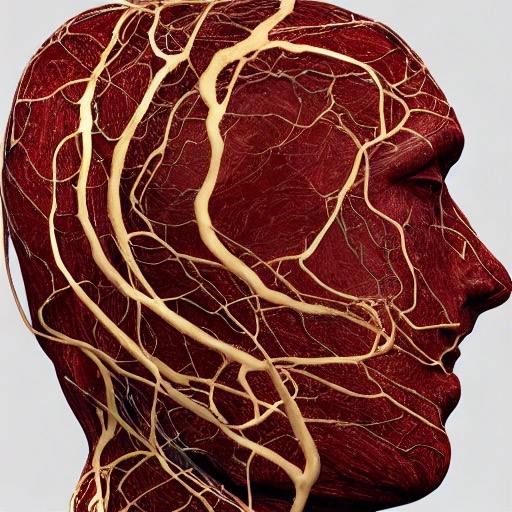} &
    \includegraphics[width=0.09\linewidth,height=0.09\linewidth]{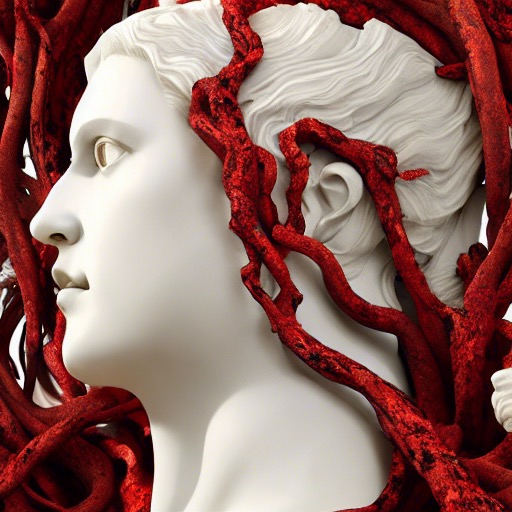} &
    \includegraphics[width=0.09\linewidth,height=0.09\linewidth]{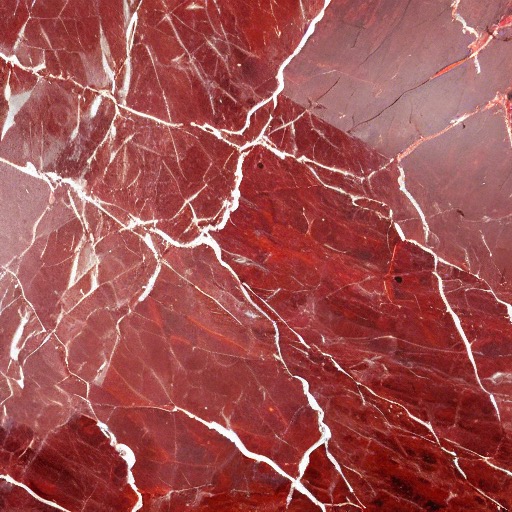} & 
    \includegraphics[width=0.09\linewidth,height=0.09\linewidth]{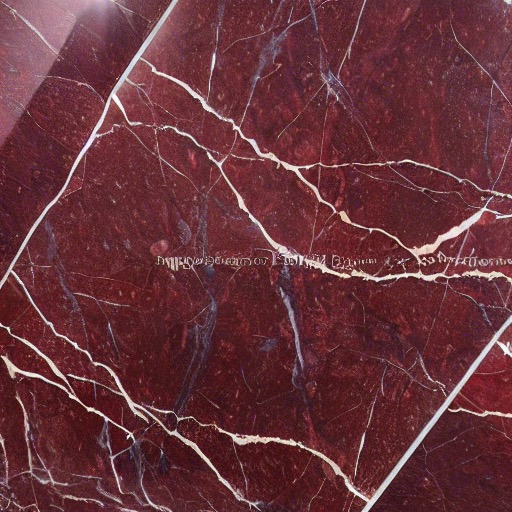} \\[2pt]
    
    \raisebox{0.045\linewidth}{\footnotesize\begin{tabular}{c@{}c@{}c@{}c@{}} Personalized \\ with $\mbox{SAC}_{8+}$ \end{tabular}}  &
    \includegraphics[width=0.09\linewidth,height=0.09\linewidth]{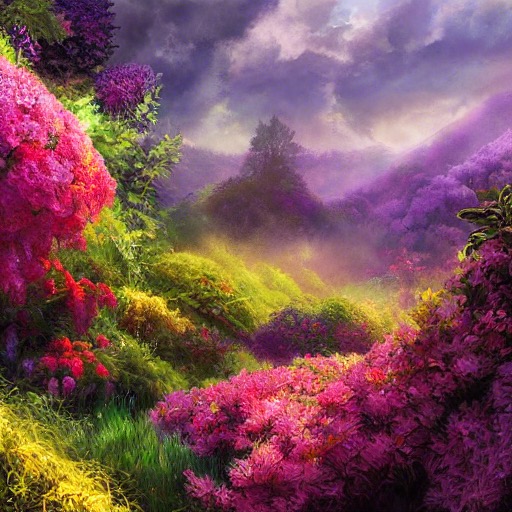} & 
    \includegraphics[width=0.09\linewidth,height=0.09\linewidth]{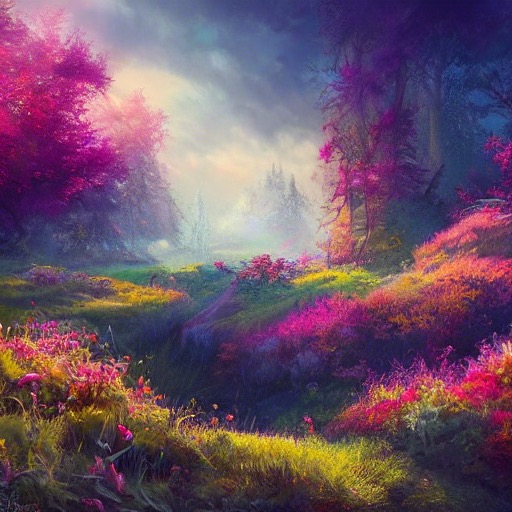} & 
    \includegraphics[width=0.09\linewidth,height=0.09\linewidth]{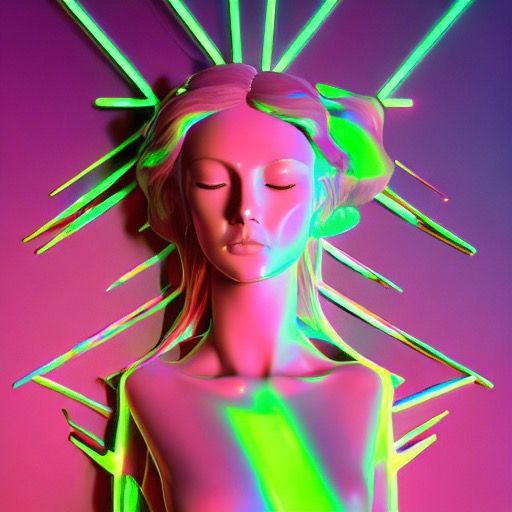} & 
    \includegraphics[width=0.09\linewidth,height=0.09\linewidth]{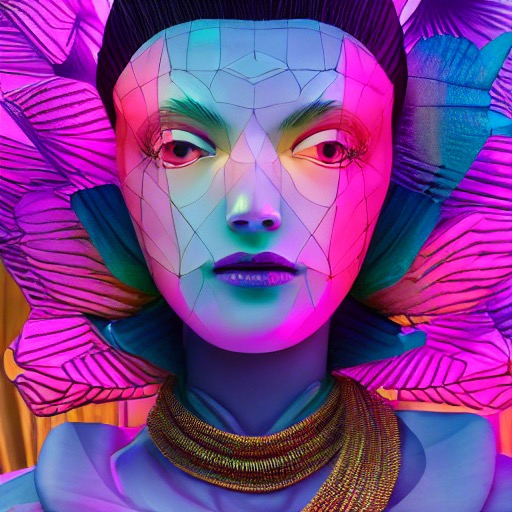} & 
    \includegraphics[width=0.09\linewidth,height=0.09\linewidth]{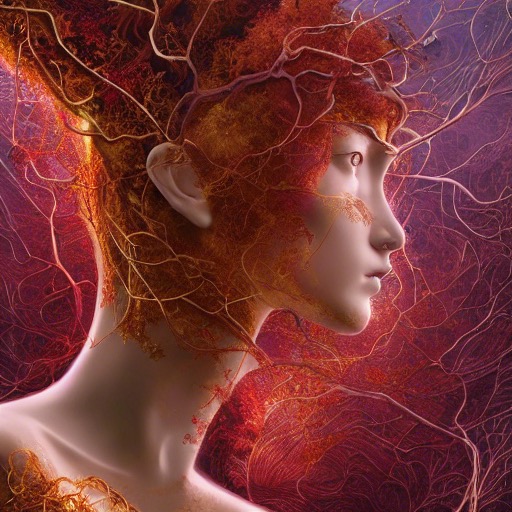} &
    \includegraphics[width=0.09\linewidth,height=0.09\linewidth]{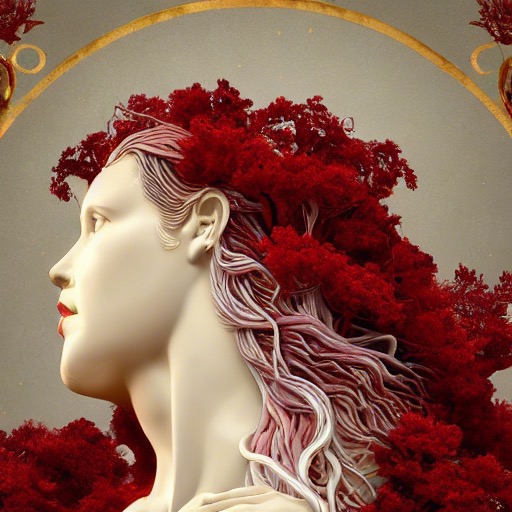} &
    \includegraphics[width=0.09\linewidth,height=0.09\linewidth]{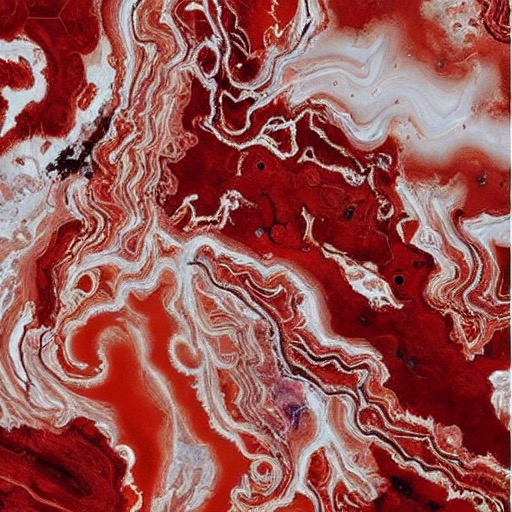} & 
    \includegraphics[width=0.09\linewidth,height=0.09\linewidth]{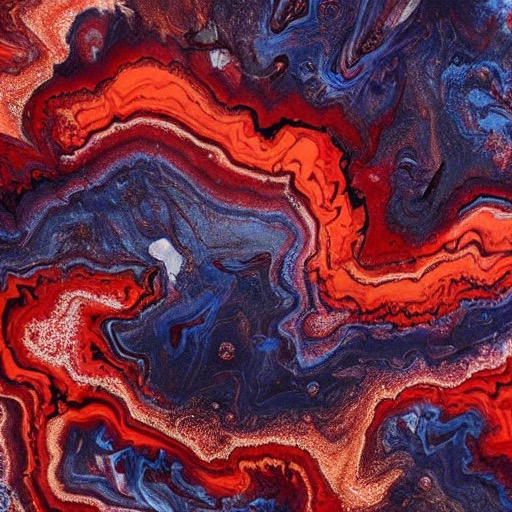} \\[-4pt]
        &
    \multicolumn{2}{l}{\tiny\begin{tabular}{c@{}c@{}c@{}} A clearing filled with colorful\\ plants in a thick wood where time\\ has stopped, trending on Artstation \end{tabular}} &
    \multicolumn{2}{l}{\tiny\begin{tabular}{c@{}c@{}c@{}} A beautiful mannequin made\\ of marble printed in 3d \\geometric neon + kintsugi... \end{tabular}} & 
    \multicolumn{2}{l}{\tiny\begin{tabular}{c@{}c@{}c@{}} Photorealistic white marble greek \\ goddess face profile sculpture \\ entwined by crimson vines... \end{tabular}} & 
    \multicolumn{2}{l}{\tiny\begin{tabular}{c@{}c@{}c@{}} Marble Polished Tile. \\Dark Volcano is an impressive\\ dark red quartzite... \end{tabular}} \\[3pt]

     \raisebox{0.045\linewidth}{\footnotesize\begin{tabular}{c@{}c@{}} Original \\ SD \end{tabular}} & 
    \includegraphics[width=0.09\linewidth,height=0.09\linewidth]{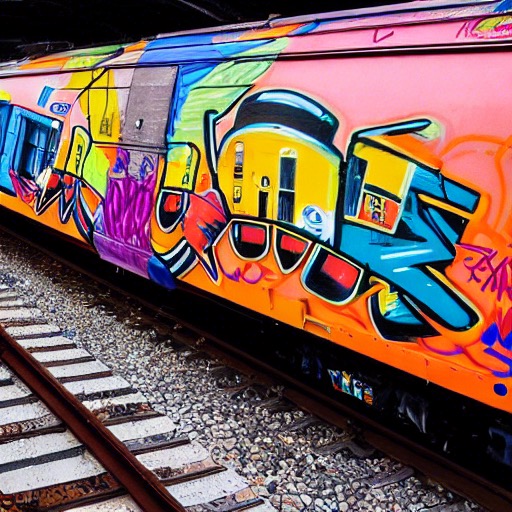} & 
    \includegraphics[width=0.09\linewidth,height=0.09\linewidth]{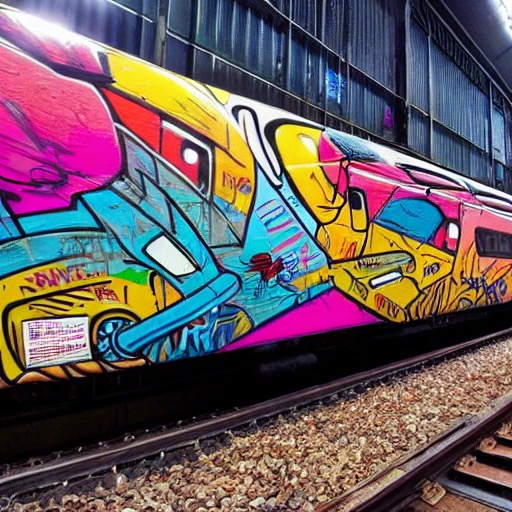} & 
    \includegraphics[width=0.09\linewidth,height=0.09\linewidth]{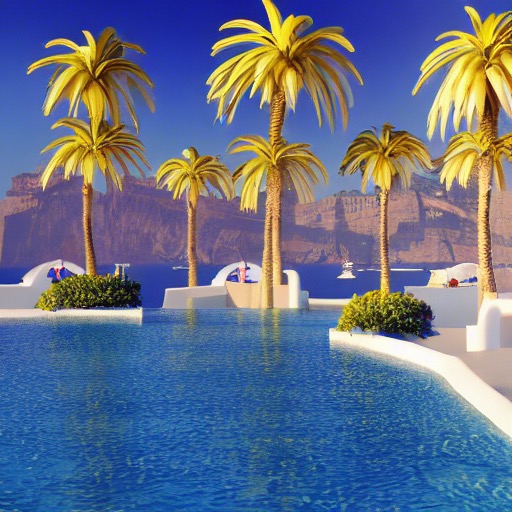} & 
    \includegraphics[width=0.09\linewidth,height=0.09\linewidth]{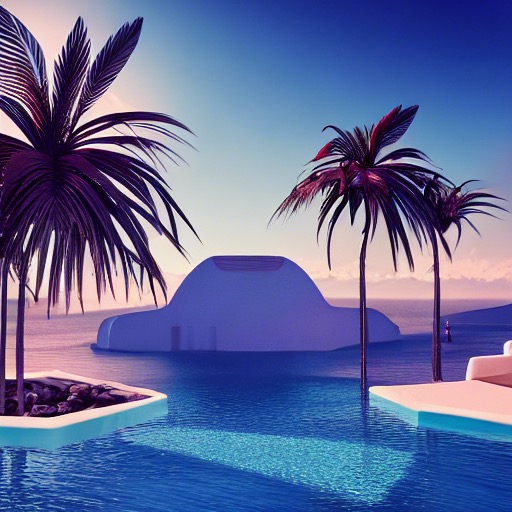} & 
    \includegraphics[width=0.09\linewidth,height=0.09\linewidth]{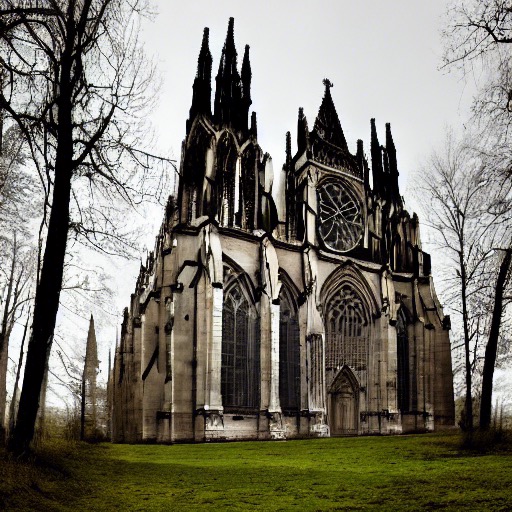} &
    \includegraphics[width=0.09\linewidth,height=0.09\linewidth]{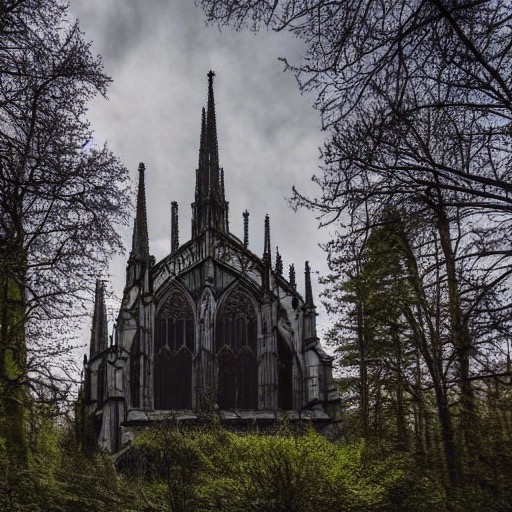} &
    \includegraphics[width=0.09\linewidth,height=0.09\linewidth]{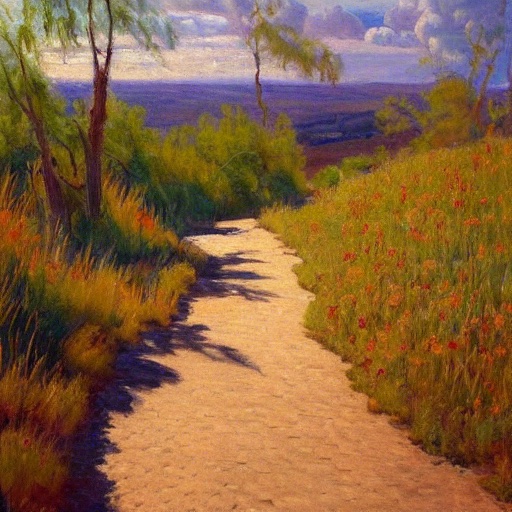} & 
    \includegraphics[width=0.09\linewidth,height=0.09\linewidth]{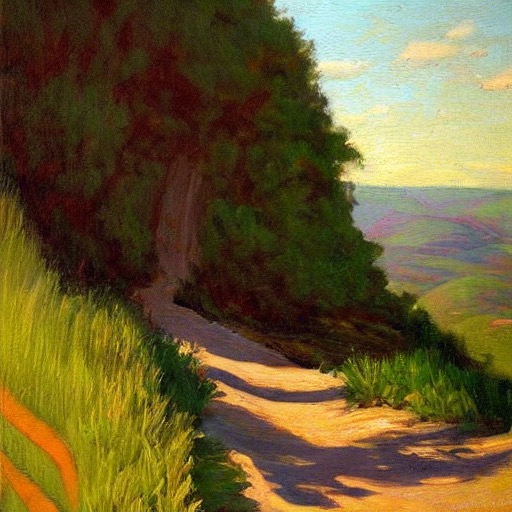} \\[2pt]
    
    \raisebox{0.045\linewidth}{\footnotesize\begin{tabular}{c@{}c@{}c@{}c@{}} Personalized \\ with $\mbox{LAION}_{7+}$ \end{tabular}}  &
    \includegraphics[width=0.09\linewidth,height=0.09\linewidth]{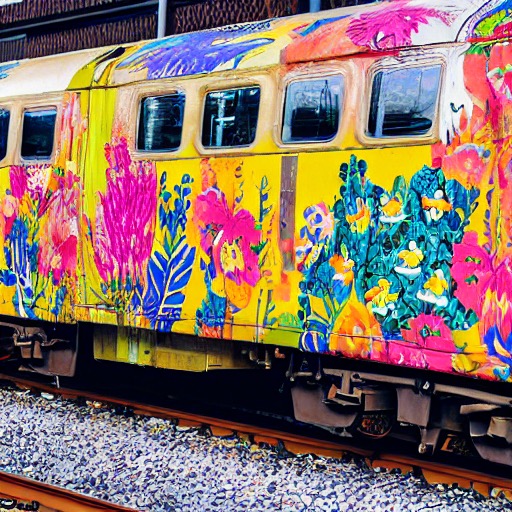} & 
    \includegraphics[width=0.09\linewidth,height=0.09\linewidth]{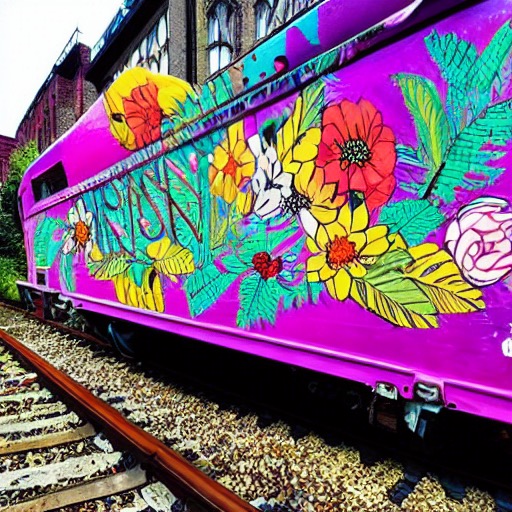} & 
    \includegraphics[width=0.09\linewidth,height=0.09\linewidth]{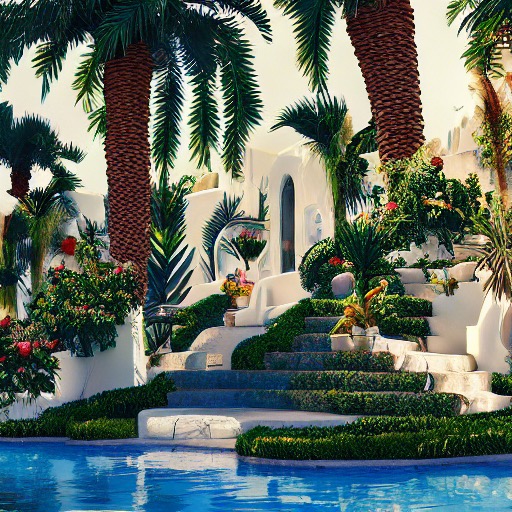} & 
    \includegraphics[width=0.09\linewidth,height=0.09\linewidth]{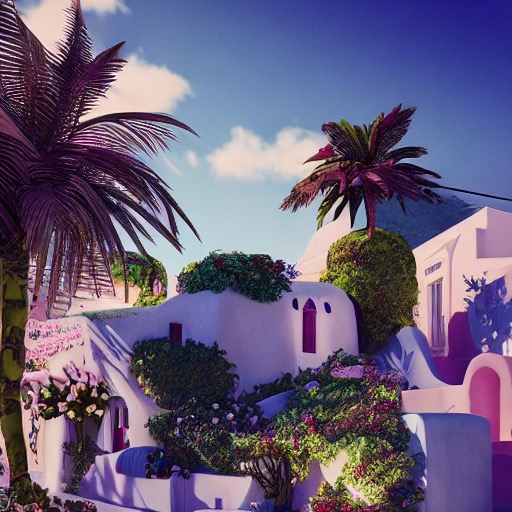} & 
    \includegraphics[width=0.09\linewidth,height=0.09\linewidth]{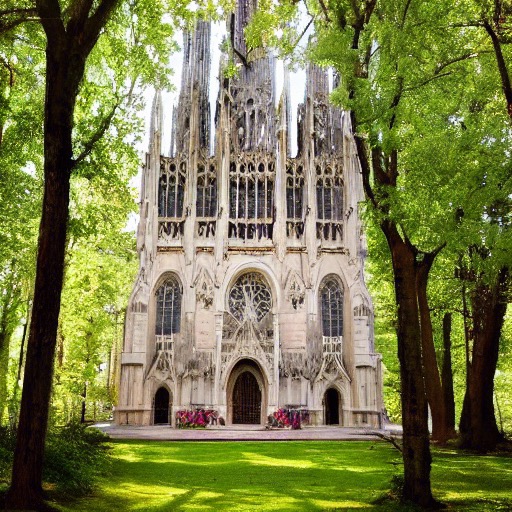} &
    \includegraphics[width=0.09\linewidth,height=0.09\linewidth]{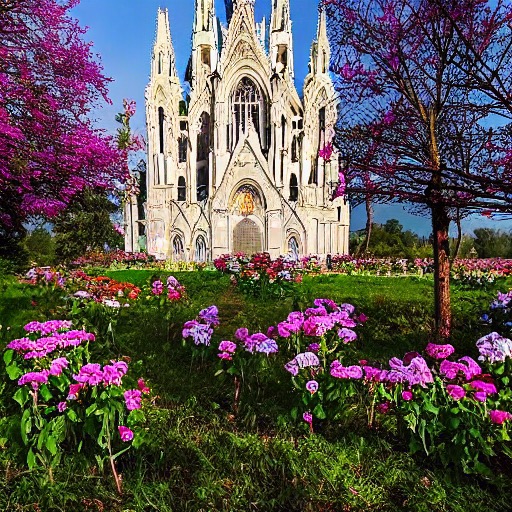} &
    \includegraphics[width=0.09\linewidth,height=0.09\linewidth]{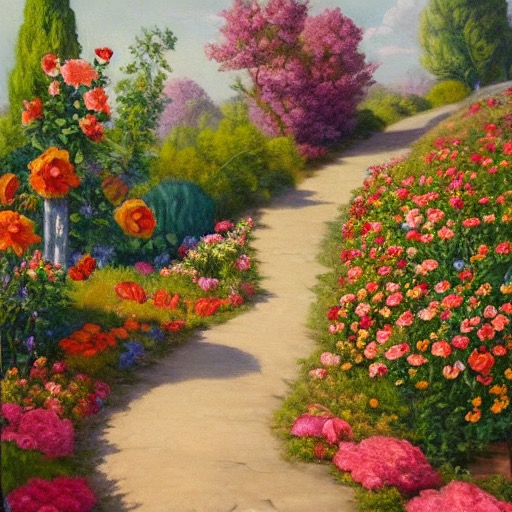} & 
    \includegraphics[width=0.09\linewidth,height=0.09\linewidth]{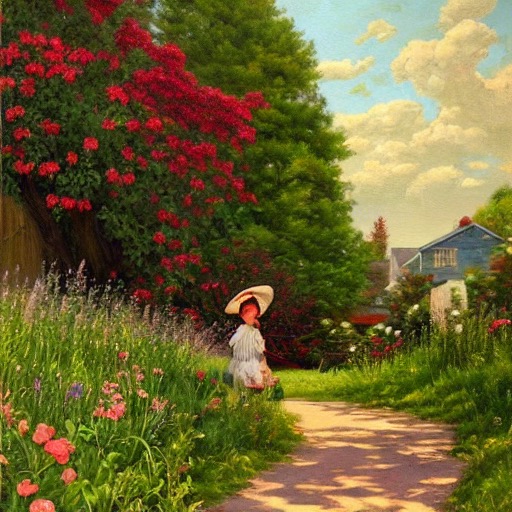} \\[-4pt]
        &
    \multicolumn{2}{l}{\tiny\begin{tabular}{c@{}c@{}} A whole train covered in wildstyle \\graffiti in the style of Seen, HQ\\ masterpiece, vibrant \end{tabular}} &
    \multicolumn{2}{l}{\tiny\begin{tabular}{c@{}c@{}c@{}} Art Deco in Santorini island \\palm trees crystal clear neon water,\\ trending Artstation, octane render... \end{tabular}} & 
    \multicolumn{2}{l}{\tiny\begin{tabular}{c@{}c@{}c@{}} A gothic cathedral in the\\ middle of the forest \end{tabular}} & 
    \multicolumn{2}{l}{\tiny\begin{tabular}{c@{}c@{}c@{}} A painting of a path  going up \\a hill, oil on canvas \end{tabular}} \\[3pt]

    \end{tabular}
    
    \end{tabular}
    
    }
    \caption{Stable diffusion generations for the original model and personalized variants using $\mbox{SAC}_{8+}$ and $\mbox{LAION}_{7+}$ aesthetic embeddings.}
    \label{fig:user_sentences_comp} 

\end{figure}
\label{fig:one}
\paragraph{Introduction}
Recently developed text-to-image models have demonstrated an unprecedented capability to reason over natural language descriptions \cite{latentdiffusion}. Their use, however, is constrained by the user’s ability to describe the desired target through text. But can generative models really grasp the aesthetic vibe preferred by its users?
Recent works such as \emph{textual inversion} \cite{gal2022textual} and \emph{dreambooth} \cite{ruiz2022dreambooth} aim to provide user personalization to diffusion models, but rather focus on learning custom objects from few images. Instead, in this work we present an alternative approach for \emph{personalization} of text-to-image diffusion models. Our goal is to guide the generative process towards custom aesthetics defined by the user, without restricting us to single objects, but rather aesthetic patterns specified by arbitrarily large sets of images. 

\paragraph{Method} At first, the user chooses a textual prompt $y$ to guide the generation. This prompt is passed through the CLIP text encoder \cite{clip} to obtain a textual embedding $c = \mbox{CLIP}_{\theta, txt}(y)$. Conditioned on this representation $c$, the diffusion process generates a image that matches the prompt. 
We propose to modify the previous representation $c$ by taking into account another embedding, representing the aesthetic preferences of the user.
Let $\lbrace x_i \rbrace_{i=1}^K$ be a set of $K$ images representing the aesthetic preference of an user. We define its corresponding \emph{aesthetic embedding} $e$ as the average of the visual embeddings of the previous images, that is, $e = \frac{1}{K}\sum_{i=1}^K \mbox{CLIP}_{\theta, vis} (x_i)$. Then we normalize the resulting vector to be of unitary norm. 
The similarity between the two embeddings, computed as the dot product $c e^{\intercal}$, can be used to measure the agreement between CLIP representation of the textual prompt and the preferences of the user. Thus, the previous expression can be used as a loss and we can perform gradient descent with respect to CLIP text encoder weights to drive the prompt representation towards the aesthetics of the user: $\theta' = \theta + \epsilon\, \nabla_{\theta} \mbox{CLIP}_{\theta, txt}(y) e^{\intercal}$, with $\epsilon$ being a user-defined step size. After a few iterations, we compute the new, personalized prompt representation, $c' = \mbox{CLIP}_{\theta', txt}(y)$, and the generation continues using the underlying diffusion process. The resulting representation $c'$ is more aligned to the user preference, while preserving the original semantics, as we will see in the experiments from the next section. Note that only the weights of the CLIP text encoder are modified, nor the visual encoder nor any other component of the diffusion model.

The benefits of our approach to personalization are several: (i) it works agnostically of the diffusion model, that is, it only requires a diffusion model which conditions on a textual prompt processed by CLIP. In the experiments we use the recent \emph{stable diffusion} (SD) model. (ii) it is computationally cheap, as it only requires a few gradient steps (less than 20 in the experiments) of the CLIP text encoder. It is not necessary to fine-tune the diffusion model, thus making it GPU-friendly. (iii) the user only needs to store one aesthetic embedding per set of images, thus saving storage space and being amenable to sharing. In the case of CLIP-L/14, the variant used by SD, $e$ is a vector of 768 dimensions. A potential drawback of our approach is that we introduce two new hyperparameters: the step size $\epsilon$ and the number of iterations. For our experiments, we set $\epsilon=1e-4$ and vary the number of iterations from 5 to 20; yet these are two levers that the user can control.
\paragraph{Experiments and results}\label{sec:exp}
First, we qualitatively show the effectiveness of our aesthetic gradients approach using several aesthetic embeddings and a collection of prompts of varying length and complexity. As the aesthetic embeddings, for this experiment we use two sets of images: $\mbox{SAC}_{8+}$, a subset of Simulacra Aesthetic Captions \cite{pressmancrowson2022} with images filtered to have a rating greater than 8; and $\mbox{LAION}_{7+}$: a subset of LAION Aesthetics v1 \cite{laionaesthetics} with images filtered to have a rating greater than 7. Figure 1 shows several generations, comparing the original SD model with the personalizations. Note that the personalized generations better reflect the aesthetics of each embedding, such as more fantasy-like imagery in the case of $\mbox{SAC}_{8+}$, and more floral patterns in $\mbox{LAION}_{7+}$.
\begin{wrapfigure}{r}{0.2\textwidth}
    \centering
    \includegraphics[width=0.2\textwidth]{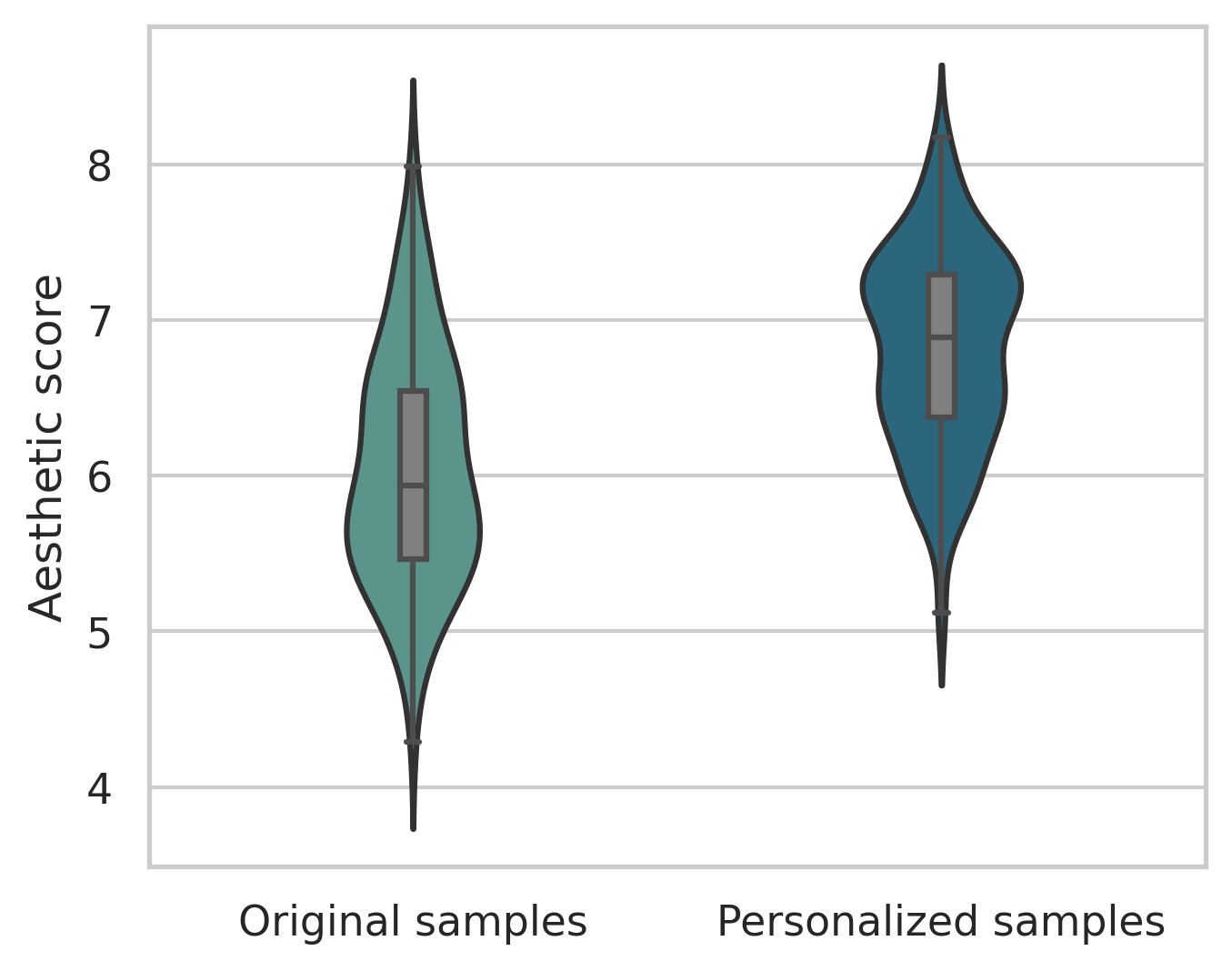}\label{fig:scores}
    \vspace{-21pt}
    \caption{Aesthetic scores distribution}
\end{wrapfigure}
We also perform an experiment to quantitatively assess the performance of the aesthetic gradient approach. To do so, we generate a list of 25 prompts, of varying length and complexity (Table \ref{tab:prompts}). For each, we generate six images, both using the original SD model and personalizing with $\mbox{SAC}_{8+}$. Finally, for each image we compute its aesthetic score using an open source model \cite{crowson2022}. The distribution of the scores for each of the two groups of generations is shown in Figure 2. Note the personalized models has improved aesthetic score, even though the diffusion model has not been optimized with respect to it.
Appendix \ref{sec:more_exps} includes another batch of experiments, using different aesthetics.
\paragraph{Conclusion}
We have proposed a flexible and efficient approach for personalizing text-to-image models, in particular by guiding the generation towards the aesthetic preferences of the user. As further work, it is straightforward to adapt our method to CLIP-guided diffusion \cite{nichol2021glide}, in which the customized CLIP model guides the generation at every timestep of the diffusion process. Also, instead of gradient descent optimizers, we could use SG-MCMC samplers \cite{zhang2019cyclical,gallego2018stochastic} to explore a greater region of the latent space and improve the diversity of the results.

\newpage
\section*{Ethical Implications}

This work aims to provide users with an effective framework for adapting the output of diffusion models to the preferred aesthetics of the user. While general text-to-image models might be biased towards specific attributes when synthesizing images from text, our approach enables the user to better reflect the desirable effects. On the other hand, malicious agents might try to use generated images to mislead viewers. This is a common issue, existing in other generative modeling approaches. Future research in generative modeling and personalization must continue investigating and revaluing these issues.

{
\small

\bibliographystyle{plain} 
\bibliography{references}

}


\appendix
\section*{Additional experiments and results}\label{sec:more_exps}

\begin{table}
\centering
\caption{Prompts used for the quantitative experiment}\label{tab:prompts}
\texttt{
\resizebox{\columnwidth}{!}{
\begin{tabular}{l} 
\hline
 \texttt{A fountain, sculpture} \\ 
\hline
 \texttt{A pyramid over a snowy scenery} \\ 
\hline
  \texttt{A giant octopus, bioluminescence}\\ 
\hline
\texttt{A still life of flowers, volumetric lighting}  \\ 
\hline
 \texttt{A still life of flowers, stained glass} \\ 
\hline
 \texttt{A nighttime cityscape, concept art} \\ 
\hline
 \texttt{The sacred library by Simon Stålenhag and Thomas Kinkade, oil on canvas} \\ 
\hline
\texttt{A gateway between dreams}  \\ 
\hline
 \texttt{Space jellyfish, watercolor} \\ 
\hline
 \texttt{Giant skull without a lower jaw, floating above a pile of gems while}\\ \texttt{it leaks gems and bone. An orange, cloudy sky fills the background} \\ 
\hline
  \texttt{An orange overstuffed chair, custom design}\\ 
\hline
 \texttt{Ethereal} \\ 
\hline
 \texttt{A clearing filled with colorful plants in a thick woods} \\ \texttt{where time has stopped, trending on Artstation} \\ 
\hline
 \texttt{An archer lounging against a tree with petals falling,}\\ \texttt{painting by Horace Vernet} \\
\hline
 \texttt{Textless, 8k, hyperdetail Papier-mache, Ambient occlusion High key light,}\\ \texttt{ Contour rivalry, octane render redshift render, Porcelain painted}\\ \texttt{ceramics by Krystle Mitchell, The efficient panda surrounds bangle,} \\ \texttt{ascot plain peel postfix circadian sunroom}\\
\hline
Dimming dares to swifting ruins lights, charges changes on the skies\\ from above, blinks true to throughout, to a closing of hands on spacing\\ world to binding breaks of recreating strings, then to dusting fantasy\\ of hands that try to wave a reach of each, and a\\  spine of splitting reeks of falling sense of decaying skying\\
\hline
Centuries of citadels, and been tuning in tones that been crystalize in a\\ field that felt a widing in its own, still a lighten\\ abyss vision for depths, it still crystalline in souls\\ that truly enjoyed, of a meaning\\
\hline
White marble, white marble bas relief profile sculpture of a \\ beautiful black haired woman with pale skin and a crown on her head \\ sitted on an intricate metal throne, medusa, white and gold kintsugi,\\ feminine shapes, crabs, spiders, scorpions, tarantulas, stunning,\\ art by hr geiger and ridley scott and alphonse mucha\\ and josephine wall, highly detailed, intricately detailed\\
\hline
Photorealistic white marble greek goddess face profile sculpture\\ entwined by golden and crimson vines and roots, flesh shows at\\ some parts under the broken marble, swirling liquified meat and \\red kintsugi, symbolist, visionary, etheric, entwined with iridiscent\\ fractal lace, alien botanicals, cinematic composition, cinematic lighting\\
\hline
A beautiful mannequin made of marble printed in 3 d geometric neon\\ + kintsugi, facing a giant doorway opening with a neon pink light,\\ flowering iridescent pineapples + orchids, transcendent, vibrant color,\\ clean linework, finely detailed, 4k, trending on artstation,\\ photorealistic, volumetric lighting, octane render \\
\hline
A pirate ship, sepia coloring, hyper-detailed, dusk, 4k octane render \\
\hline
Vaporwave soviet skyline at sunrise, trending on Artstation. \\Many intrincate details\\
\hline
Marble Polished Tile. Sky Blue is an impressive pale blue quartzite.\\ Its appearance is reminiscent of a splendid blue sky interspersed with\\ fluffy white clouds, as its name suggests. This natural stone's\\ base shuffles different soft blues such as blue lavender,\\ pale blue, and pastel indigo. The veins look like clouds. \\Decorative marble tile\\
\hline
A photograph of an astronaut riding a horse \\
\hline
A painting of a tree, oil on canvas
\\ \hline
\end{tabular}}
}

\end{table}

Figure 3 depicts further qualitative experiments, with different aesthetics of those from the main text. In particular, we target the following aesthetic preferences: \texttt{Aivazovsky} (five paintings from the artist),
\texttt{cloudcore}, \texttt{gloomcore}, and \texttt{glowwave}, with the last three consisting in 100 random images scraped from Pinterest using those terms as the keyword. In the Figure, the first column shows the generations for the original SD model, the second column shows the generations also for the original model but appending the aesthetic keyword at the end of the prompt; and the third column show the generations personalized with the proposed aesthetic gradients approach. Note that whereas the second column shows that the effect of appending the aesthetic keyword in the prompt has a negligible or very dim effect due to the limitations of the CLIP text encoder, the aesthetic gradients approach can better reflect the aesthetic effect that is to be expected, dramatically improving the corresponding aesthetic vibe of the generations.

\begin{figure}[!hbt]
    \centering
    \setlength{\abovecaptionskip}{6.5pt}
    \setlength{\belowcaptionskip}{-3.5pt}
    \setlength{\tabcolsep}{0.55pt}
    \renewcommand{\arraystretch}{1.0}
    {
    
    \begin{tabular}{c}
    
    \begin{tabular}{c@{\hskip 25pt} c @{\hskip 50pt} c @{\hskip 50pt} c }
     & \multicolumn{1}{l}{Original SD} & \multicolumn{1}{l}{Original SD} & \multicolumn{1}{l}{Personalized}
     \\

    \raisebox{0.09\linewidth}{\footnotesize\begin{tabular}{l@{}l@{}} \texttt{aivazovsky} \end{tabular}} & 
    \includegraphics[width=0.19\linewidth,height=0.19\linewidth]{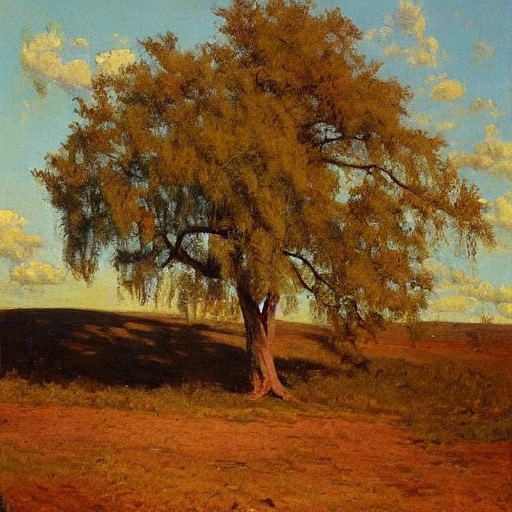} & 
    \includegraphics[width=0.19\linewidth,height=0.19\linewidth]{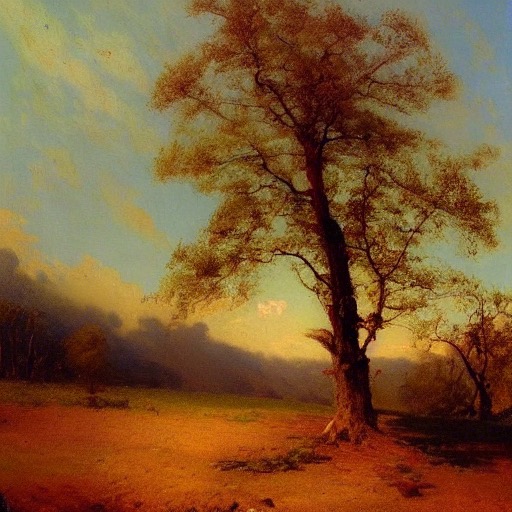} & 
    \includegraphics[width=0.19\linewidth,height=0.19\linewidth]{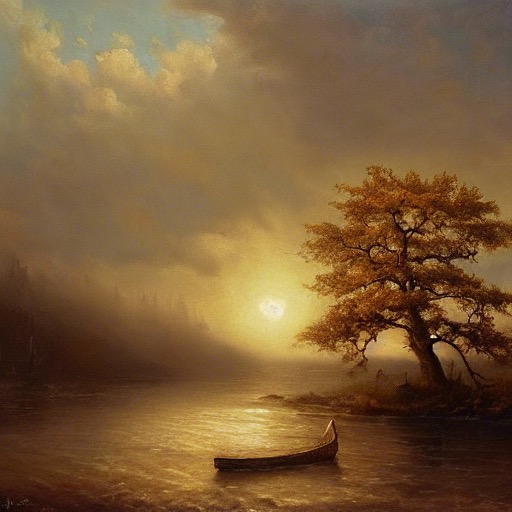} \\[-4pt]
     & 
     \multicolumn{1}{l}{\tiny\begin{tabular}{l@{}l@{}l@{}} A painting of a tree, oil on canvas \end{tabular}} &
    \multicolumn{1}{l}{\tiny\begin{tabular}{l@{}l@{}l@{}} A painting of a tree, oil on canvas\\ by Ivan Aivazovsky \end{tabular}} & 
    \multicolumn{1}{l}{\tiny\begin{tabular}{l@{}l@{}l@{}} A painting of a tree, oil on canvas \end{tabular}} \\[3pt]
    
    \raisebox{0.09\linewidth}{\footnotesize\begin{tabular}{l@{}l@{}l@{}l@{}} \texttt{cloudcore}\end{tabular}}  &
    \includegraphics[width=0.19\linewidth,height=0.19\linewidth]{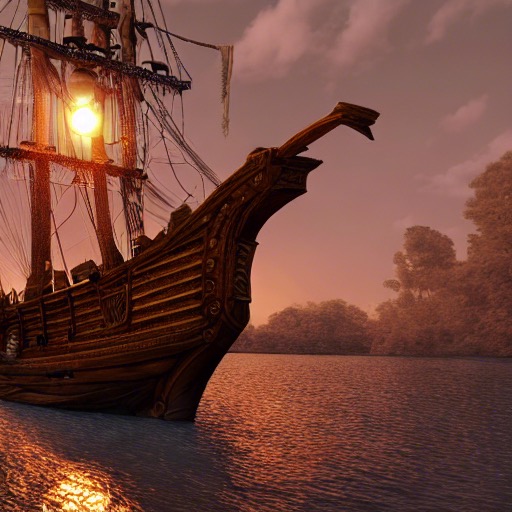} & 
    \includegraphics[width=0.19\linewidth,height=0.19\linewidth]{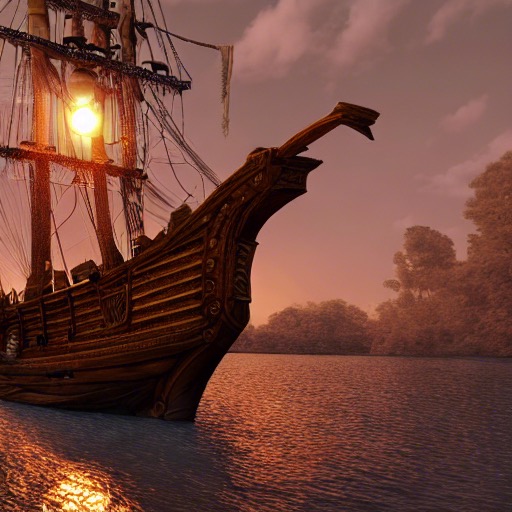} & 
    \includegraphics[width=0.19\linewidth,height=0.19\linewidth]{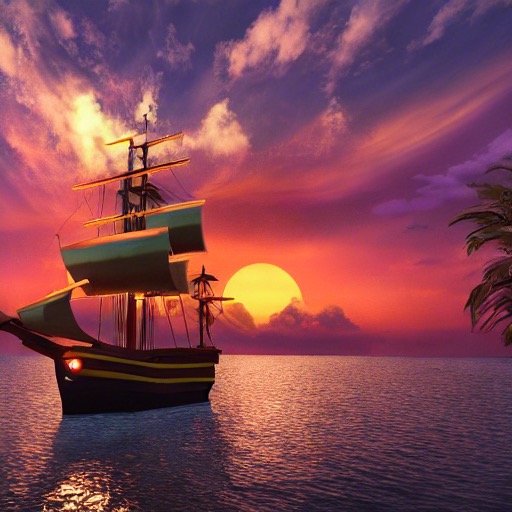} \\[-4pt]
        &
    \multicolumn{1}{l}{\tiny\begin{tabular}{l@{}l@{}l@{}} A pirate ship, sepia coloring, hyper-\\detailed, dusk, 4k octane render \end{tabular}} &
    \multicolumn{1}{l}{\tiny\begin{tabular}{l@{}l@{}l@{}} A pirate ship, sepia coloring, hyper-\\..., cloudcore \end{tabular}} & 
    \multicolumn{1}{l}{\tiny\begin{tabular}{l@{}l@{}l@{}} A pirate ship, sepia coloring, hyper-\\detailed, dusk, 4k octane render \end{tabular}} \\[5pt]
    
  \raisebox{0.09\linewidth}{\footnotesize\begin{tabular}{l@{}l@{}l@{}l@{}} \texttt{gloomcore}  \end{tabular}}  &
    \includegraphics[width=0.19\linewidth,height=0.19\linewidth]{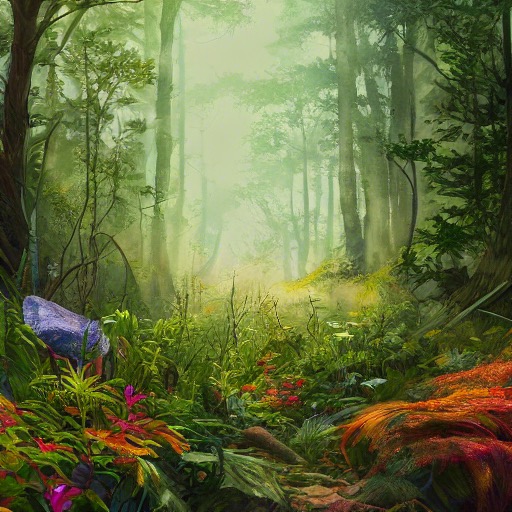} & 
    \includegraphics[width=0.19\linewidth,height=0.19\linewidth]{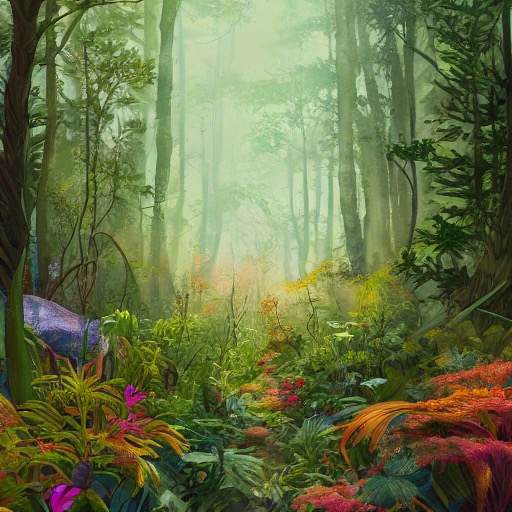} & 
    \includegraphics[width=0.19\linewidth,height=0.19\linewidth]{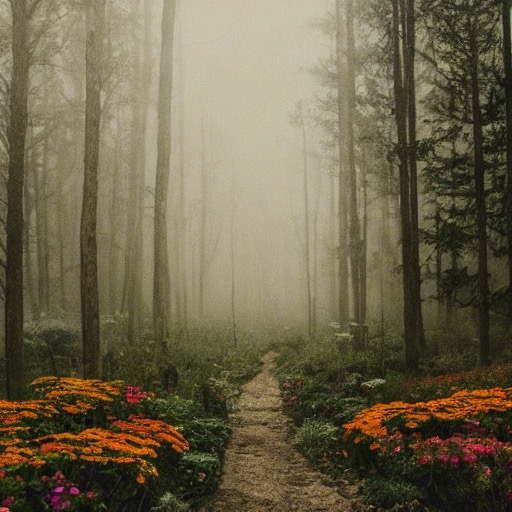} \\[-4pt]
        &
    \multicolumn{1}{l}{\tiny\begin{tabular}{l@{}l@{}l@{}} A clearing filled with colorful\\ plants in a thick wood where time\\ has stopped, trending on Artstation \end{tabular}} &
    \multicolumn{1}{l}{\tiny\begin{tabular}{l@{}l@{}l@{}} A clearing filled with colorful\\ plants in a thick wood where time\\ ..., gloomcore \end{tabular}} & 
    \multicolumn{1}{l}{\tiny\begin{tabular}{l@{}l@{}l@{}} A clearing filled with colorful\\ plants in a thick wood where time\\ has stopped, trending on Artstation \end{tabular}} \\[3pt]

     \raisebox{0.09\linewidth}{\footnotesize\begin{tabular}{l@{}l@{}l@{}l@{}} \texttt{gloomcore} \end{tabular}}  &
    \includegraphics[width=0.19\linewidth,height=0.19\linewidth]{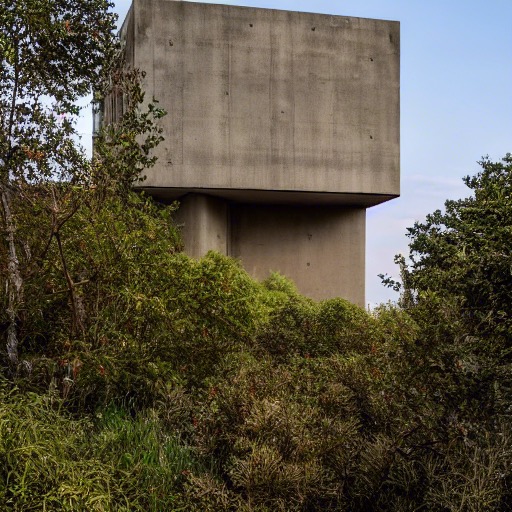} & 
    \includegraphics[width=0.19\linewidth,height=0.19\linewidth]{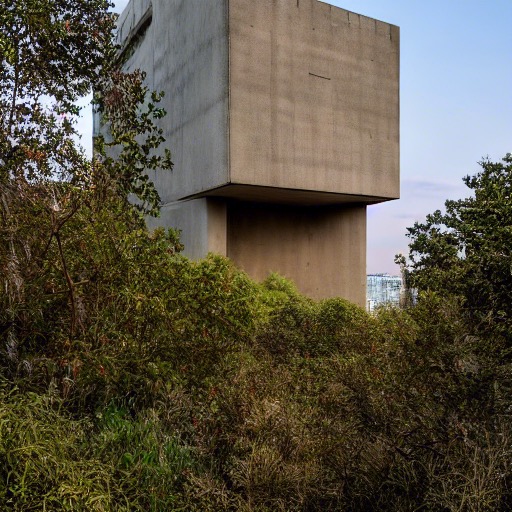} & 
    \includegraphics[width=0.19\linewidth,height=0.19\linewidth]{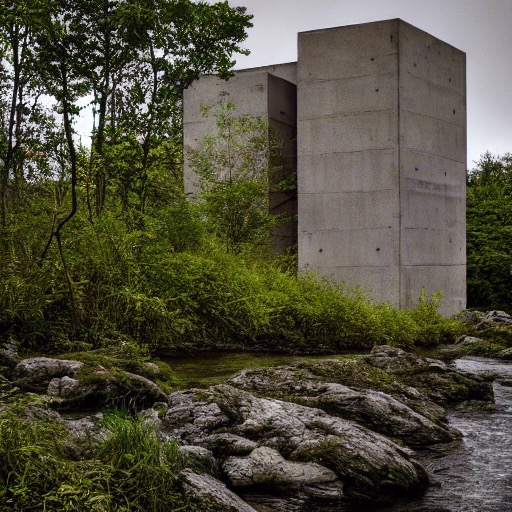} \\[-4pt]
        &
    \multicolumn{1}{l}{\tiny\begin{tabular}{l@{}l@{}l@{}} Award-wining photograph of a \\ brutalist concrete building with\\ exuberant vegetation, Provia, Velvia \end{tabular}} &
    \multicolumn{1}{l}{\tiny\begin{tabular}{l@{}l@{}l@{}} Award-wining photograph of a \\ brutalist concrete building with\\ ..., gloomcore \end{tabular}} & 
    \multicolumn{1}{l}{\tiny\begin{tabular}{l@{}l@{}l@{}} Award-wining photograph of a \\ brutalist concrete building with\\ exuberant vegetation, Provia, Velvia \end{tabular}} \\[3pt]

     \raisebox{0.09\linewidth}{\footnotesize\begin{tabular}{l@{}l@{}l@{}l@{}} \texttt{glowwave} \end{tabular}}  &
    \includegraphics[width=0.19\linewidth,height=0.19\linewidth]{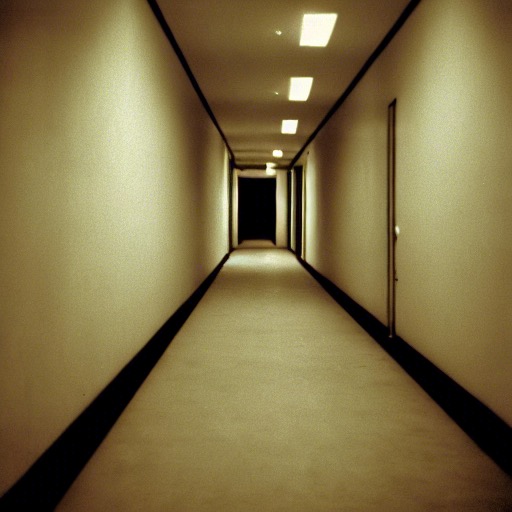} & 
    \includegraphics[width=0.19\linewidth,height=0.19\linewidth]{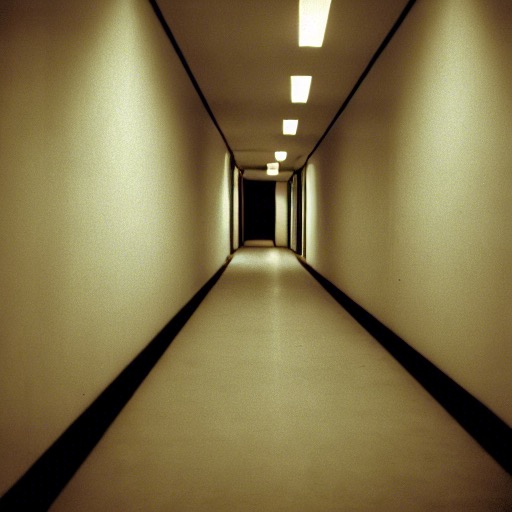} & 
    \includegraphics[width=0.19\linewidth,height=0.19\linewidth]{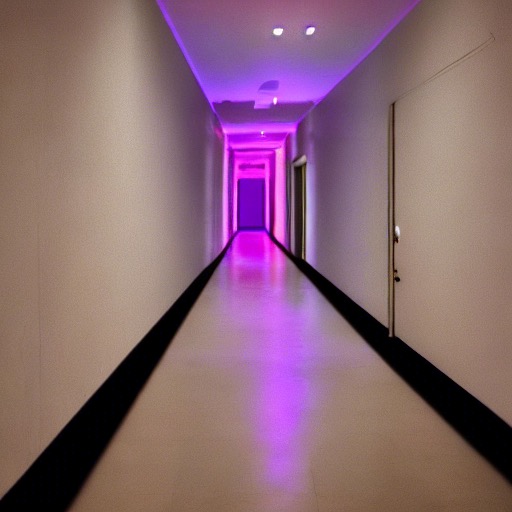} \\[-4pt]
        &
    \multicolumn{1}{l}{\tiny\begin{tabular}{l@{}l@{}l@{}} Award-wining photograph of a dark \\corridor, Provia, Velvia \end{tabular}} &
    \multicolumn{1}{l}{\tiny\begin{tabular}{l@{}l@{}l@{}} Award-wining photograph of a dark \\corridor, Provia, Velvia, glowwave \end{tabular}} & 
    \multicolumn{1}{l}{\tiny\begin{tabular}{l@{}l@{}l@{}} Award-wining photograph of a dark \\corridor, Provia, Velvia \end{tabular}} \\[3pt]

     \raisebox{0.09\linewidth}{\footnotesize\begin{tabular}{l@{}l@{}l@{}l@{}} \texttt{glowwave} \end{tabular}}  &
    \includegraphics[width=0.19\linewidth,height=0.19\linewidth]{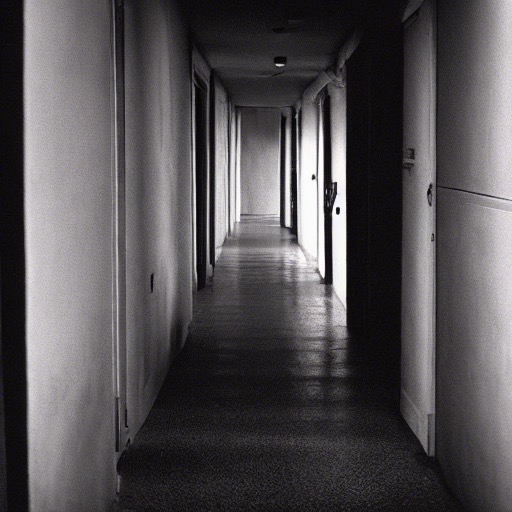} & 
    \includegraphics[width=0.19\linewidth,height=0.19\linewidth]{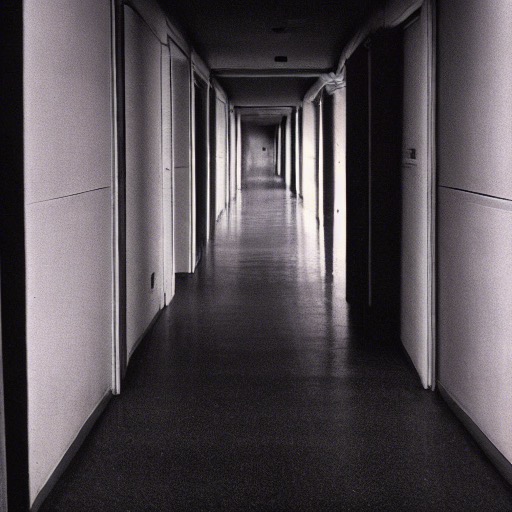} & 
    \includegraphics[width=0.19\linewidth,height=0.19\linewidth]{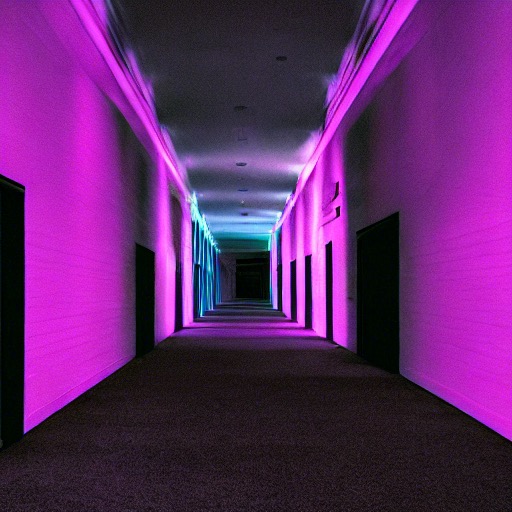} \\[-4pt]
        &
    \multicolumn{1}{l}{\tiny\begin{tabular}{l@{}l@{}l@{}} Award-wining photograph of a dark \\corridor, Provia, Velvia \end{tabular}} &
    \multicolumn{1}{l}{\tiny\begin{tabular}{l@{}l@{}l@{}} Award-wining photograph of a dark \\corridor, Provia, Velvia, glowwave \end{tabular}} & 
    \multicolumn{1}{l}{\tiny\begin{tabular}{l@{}l@{}l@{}} Award-wining photograph of a dark \\corridor, Provia, Velvia \end{tabular}} \\[3pt]

     \raisebox{0.09\linewidth}{\footnotesize\begin{tabular}{l@{}l@{}l@{}l@{}} \texttt{glowwave}  \end{tabular}}  &
    \includegraphics[width=0.19\linewidth,height=0.19\linewidth]{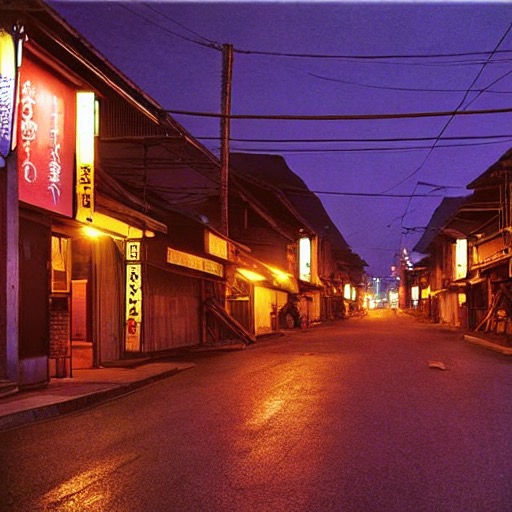} & 
    \includegraphics[width=0.19\linewidth,height=0.19\linewidth]{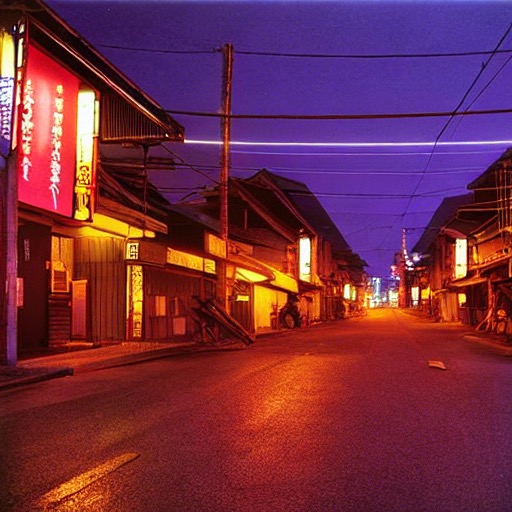} & 
    \includegraphics[width=0.19\linewidth,height=0.19\linewidth]{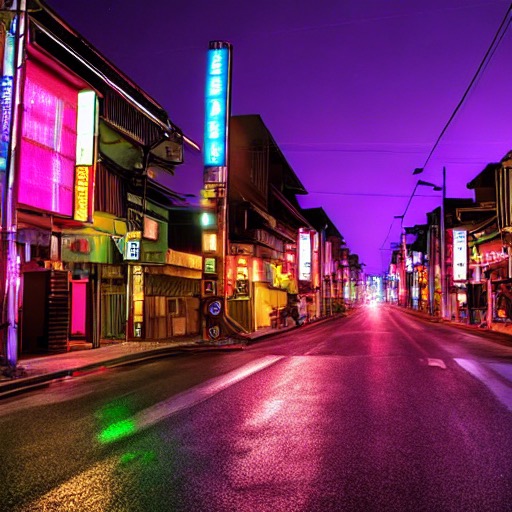} \\[-4pt]
        &
    \multicolumn{1}{l}{\tiny\begin{tabular}{l@{}l@{}l@{}} Award-wining photograph of japanese\\ town street at night, Provia, Velvia \end{tabular}} &
    \multicolumn{1}{l}{\tiny\begin{tabular}{l@{}l@{}l@{}} Award-wining photograph of japanese\\ ..., glowwave \end{tabular}} & 
    \multicolumn{1}{l}{\tiny\begin{tabular}{l@{}l@{}l@{}} Award-wining photograph of japanese\\ town street at night, Provia, Velvia \end{tabular}} \\[3pt]

    \end{tabular}
    
    \end{tabular}
    
    }
    \caption{Further qualitative results using different aesthetic embeddings.}
    \label{fig:user_sentences_comp} 

\end{figure}

\end{document}